\documentclass[lettersize,journal]{IEEEtran}
\usepackage{amsmath,amsfonts}
\usepackage{algorithmic}
\usepackage{algorithm}
\usepackage{array}
\usepackage{caption}
\usepackage[caption=false,font=normalsize,labelfont=sf,textfont=sf]{subfig}
\usepackage{textcomp}
\usepackage{stfloats}
\usepackage{url}
\usepackage{verbatim}
\usepackage{graphicx}
\usepackage{cite}
\usepackage[margin=1in]{geometry} 
\usepackage{tabularx} 
\usepackage{booktabs} 
\begin{document}

\title{Extensive Exploration in Complex Traffic Scenarios using Hierarchical Reinforcement Learning}

\author{Zhihao Zhang, Ekim Yurtsever~\IEEEmembership{Member,~IEEE}, Keith A. Redmill~\IEEEmembership{Senior Member,~IEEE}
\thanks{Manuscript received July 1, 2024; revised tbd; accepted
tbd. Date of publication tbd; date of current version tbd.
This work was supported in part by Carnegie Mellon University’s Safety21 National University Transportation Center, which is sponsored by the US Department of Transportation under grants 69A3552344811/69A3552348316.}
\thanks{The authors are with the Department of Electrical and Computer Engineering, The Ohio State University, Columbus, OH 43210 USA
(e-mail: zhang.11606@osu.edu; yurtsever.2@osu.edu; redmill.1@osu.edu).}}%
\markboth{}%
{right page header for two sided}%


\maketitle

\begin{abstract}
Developing an automated driving system capable of navigating complex traffic environments remains a formidable challenge. Unlike rule-based or supervised learning-based methods, Deep Reinforcement Learning (DRL) based controllers eliminate the need for domain-specific knowledge and datasets, thus providing adaptability to various scenarios. Nonetheless, a common limitation of existing studies on DRL-based controllers is their focus on driving scenarios with simple traffic patterns, which hinders their capability to effectively handle complex driving environments with delayed, long-term rewards, thus compromising the generalizability of their findings.
In response to these limitations, our research introduces a pioneering hierarchical framework that efficiently decomposes intricate decision-making problems into manageable and interpretable subtasks. We adopt a two step training process that trains the high-level controller and low-level controller separately. The high-level controller exhibits an enhanced exploration potential with long-term delayed rewards, and the low-level controller provides longitudinal and lateral control ability using short-term instantaneous rewards. Through simulation experiments, we demonstrate the superiority of our hierarchical controller in managing complex highway driving situations.

\end{abstract}


\section{Introduction}
\IEEEPARstart{T}{he} emergence of autonomous vehicles has introduced an innovative trajectory in the narrative of highway driving. Autonomous highway driving, as studied in this paper, pertains to the decision-making and vehicle control processes through which autonomous vehicles navigate and interact within highway environments. It represents the convergence of cutting edge technologies and involves the integration of advanced artificial intelligent and control algorithms, sensors and sensor fusion, and real-time data processing to enable vehicles to make informed choices, such as lane changes, merging, overtaking, and responding to intricate and dynamic traffic conditions\cite{kiran2021deep,yurtsever2020survey}, promising enhancements in traffic management, safety, and overall efficiency\cite{schubert2011evaluating}.
The goal is to ensure safe, efficient, and reliable autonomous travel on highways by enabling vehicles to analyze complex scenarios, anticipate potential hazards, and execute actions that align with traffic rules and user preferences\cite{varaiya1991sketch,ngai2011multiple,xu2018reinforcement,schubert2011evaluating}. 

Yet, despite remarkable strides, the task of crafting autonomous vehicles with the capability to make discerning decisions remains a formidable obstacle. This complexity emanates from the complicated amalgamation of diverse disciplines and the intricacies of real-world scenarios, necessitating profound innovation and interdisciplinary collaboration\cite{schubert2011evaluating,xu2018reinforcement,rosenzweig2015review,peng2022integrated}.\IEEEpubidadjcol

In the rapidly evolving field of automated driving systems, three fundamental methodologies have emerged as essential strategies for addressing the complex challenge of decision-making\cite{xu2018reinforcement,peng2022integrated}.The first approach, characterized as conventional, embraces a modular framework that governs the decision-making process within highway scenarios\cite{bevly2016lane,hatipoglu2003automated,chandler1958traffic,gipps1981behavioural,treiber2000congested,kesting2007general,yurtsever2020survey}. Within this paradigm, distinct modules are meticulously crafted, each responsible for addressing specific aspects of decision-making. These modules encompass specific behaviors such as lane changes, merging, overtaking, and other maneuvers crucial for highway navigation. Ultimately, the amalgamation of these modules along with a behavior planning and selection algorithm culminates in a cohesive decision-making system, which then interfaces with vehicle control mechanisms. This approach, while promoting transparency and interpretability in decision processes, demands the meticulous calibration of diverse driving factors. It necessitates a comprehensive understanding of adapting to an array of driving scenarios while accounting for diverse safety considerations.\IEEEpubidadjcol

\begin{figure*}[!t]
\centering
\includegraphics[width=6.4in]{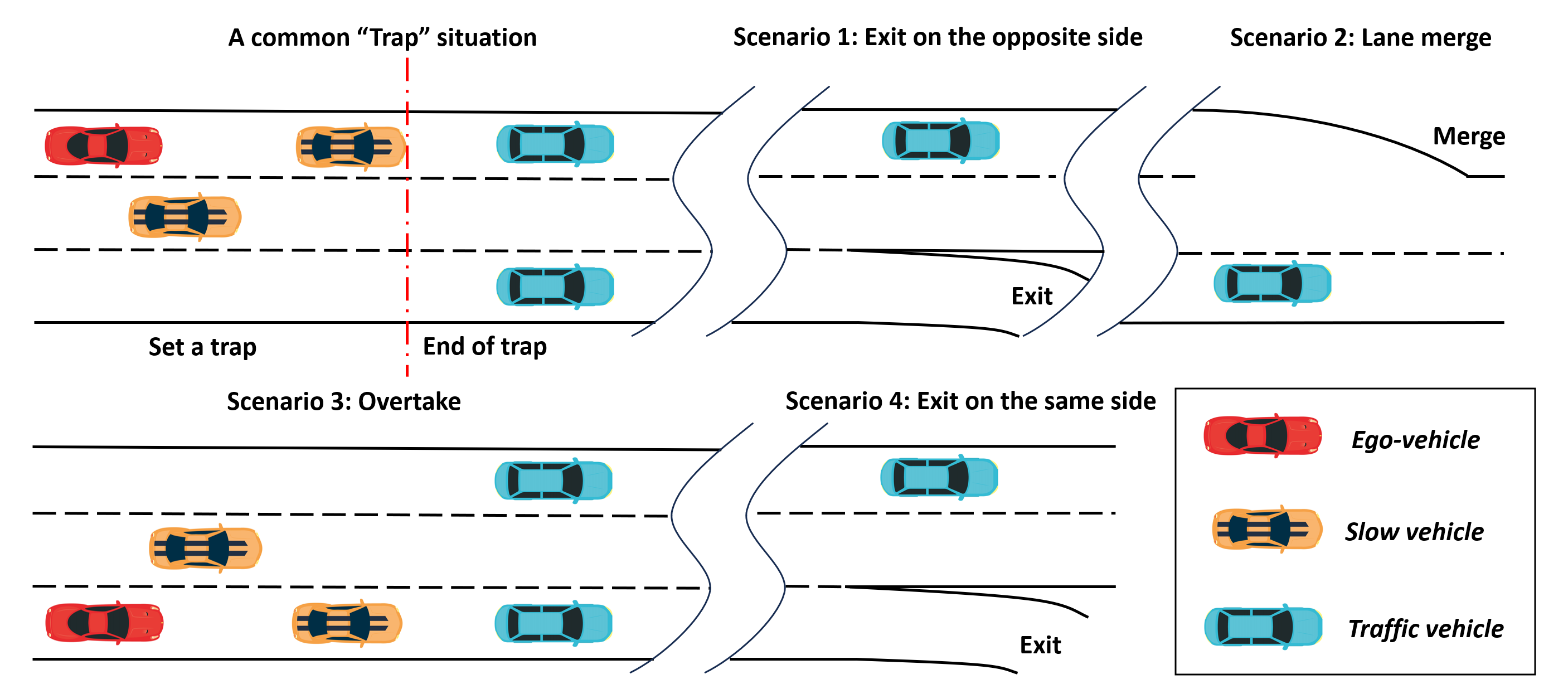}
\caption{Experiment Setting to Evaluate Exploration Ability: In this setup, the DRL agent initially encounters a group of two slow-moving traffic vehicles. This scenario tests the agent's ability to navigate through this 'trap'. Upon successfully maneuvering out of this situation, the agent is then introduced to normal traffic conditions, further assessing its adaptability and exploration capabilities.}
\label{trap}
\end{figure*}

Conversely, data-driven methodologies, exemplified by supervised learning, take a different path towards decision-making refinement\cite{han2019driving,bojarski2016end,bojarski2017explaining,codevilla2018end,hawke2020urban}. This approach simplifies the process by using machine learning methods to replicate expert driving behavior. Developing such a controller entails concurrently gathering and correlating sensor data with the respective steering, brake, and throttle actuator actions performed by human drivers. However, a notable caveat of this method is its demand for extensive data collection.

Finally, reinforcement learning emerges as another compelling paradigm, aiming to equip autonomous systems with decision-making capabilities\cite{yurtsever2020integrating,kendall2019learning,kiran2021deep}. Through simulated interactions with the environment, the self-driving agents accumulate data and experience, facilitating the acquisition of driving strategies. However, DRL controllers often rely on exploring the environment in the early stages of their training. Insufficient exploration might hinder the DRL controller from discovering better decision-making behaviors, which are essential for coping with more complex traffic environments\cite{bevly2016lane}.
Moreover, existing studies gravitate towards simpler traffic scenarios, restricting the generalizability of their findings when facing more intricate and unpredictable real-world conditions\cite{chen2018deep,sonu2018exploiting}.

To address these limitations, we introduce a novel hierarchical framework that can decompose complex highway decision-making problems and empower the driving agent with the ability to effectively navigate intricate traffic environments. The design of the two-level controller also facilitates exploration at various scales. The high-level controller employs high-level actions for broader exploration, while the lower-level controller uses low-level actions for precise vehicle control and localized exploration. 
To ensure the high-level controller's extensive exploration of the environment, we train the high-level and low-level controllers separately. The high-level controller is trained using a critic function specifically designed for highway driving environments.

To test the feasibility of this hierarchical DRL controller, we have devised a highway driving scenario that includes a 'trap' situation, as shown in Figure~\ref{trap}. This 'trap' situation is represented by two vehicles moving slower than the typical traffic, forcing the ego vehicle to reduce speed to avoid a collision. The agent's objective is to maneuver out of this "Trap" formed by slow-moving traffics and subsequently accelerate to the ideal cruising speed. 

Vehicles may encounter the necessity to navigate around slow-moving traffic under the following four scenarios:
\begin{enumerate}
\item The ego vehicle is initially positioned in a lane far from the exit it intends to take, requiring the vehicle to cross multiple lanes toward the opposite side of the highway to reach the exit.  
\item The ego vehicle is in a left lane that is merging into the main traffic flow on the right. Escape from the ”trap” allows the ego vehicle to integrate into the denser traffic without disrupting the flow or causing safety concerns. 
\item The ego vehicle, positioned in the right lane, desires to use the far-left overtaking lane to pass and move beyond the slower vehicle. 
\item The ego vehicle is in a lane close to an exit while slow-moving vehicles intend to slow down and merge to the exit, but the ego vehicle needs to continue on the main road. Escape from the ”trap” will prevent the ego vehicle from following the slow-moving vehicles to the exit. 
\end{enumerate}

Our experiments and analyses are centered on the common 'trap' situation depicted in Figure~\ref{trap}. The subsequent four specific scenarios that necessitate the ego vehicle to navigate around slow-moving traffic fall outside the scope of this paper's investigation.

Specifically, our simulations of the "Trap" situation involves the ego vehicle, the nearby trap vehicles, and further upstream traffic vehicles which are governed by a rule-based controller. The traffic vehicles create a more complex and varied traffic environment for the agent to navigate after it moves past slow-moving traffic vehicles. This diversity of traffic patterns is important in demonstrating the stronger adaptability of automated driving strategies, while also preventing the model from overfitting to just one type of traffic pattern.

Our main contributions are as follows: 
\begin{itemize}
\item A novel hierarchical DRL-based controller for highway driving scenario is proposed. The high-level controller enhances the vehicle's exploration potential with long-term planning and allows it to escape from slow traffic traps. The lower-level controller undertakes the intricate task of managing the vehicle's basic control maneuvers, fostering fine-grained control over driving dynamics.
\item A two-step training process for the high-level controller and the low-level controller is designed and implemented. 
\item A speed-biased highway driving reward function is created. A trap scenario is created to test the agent’s ability to explore long-term benefits in highway driving.
\end{itemize}
\begin{figure*}[!t]
\centering
\includegraphics[width=6.5in]{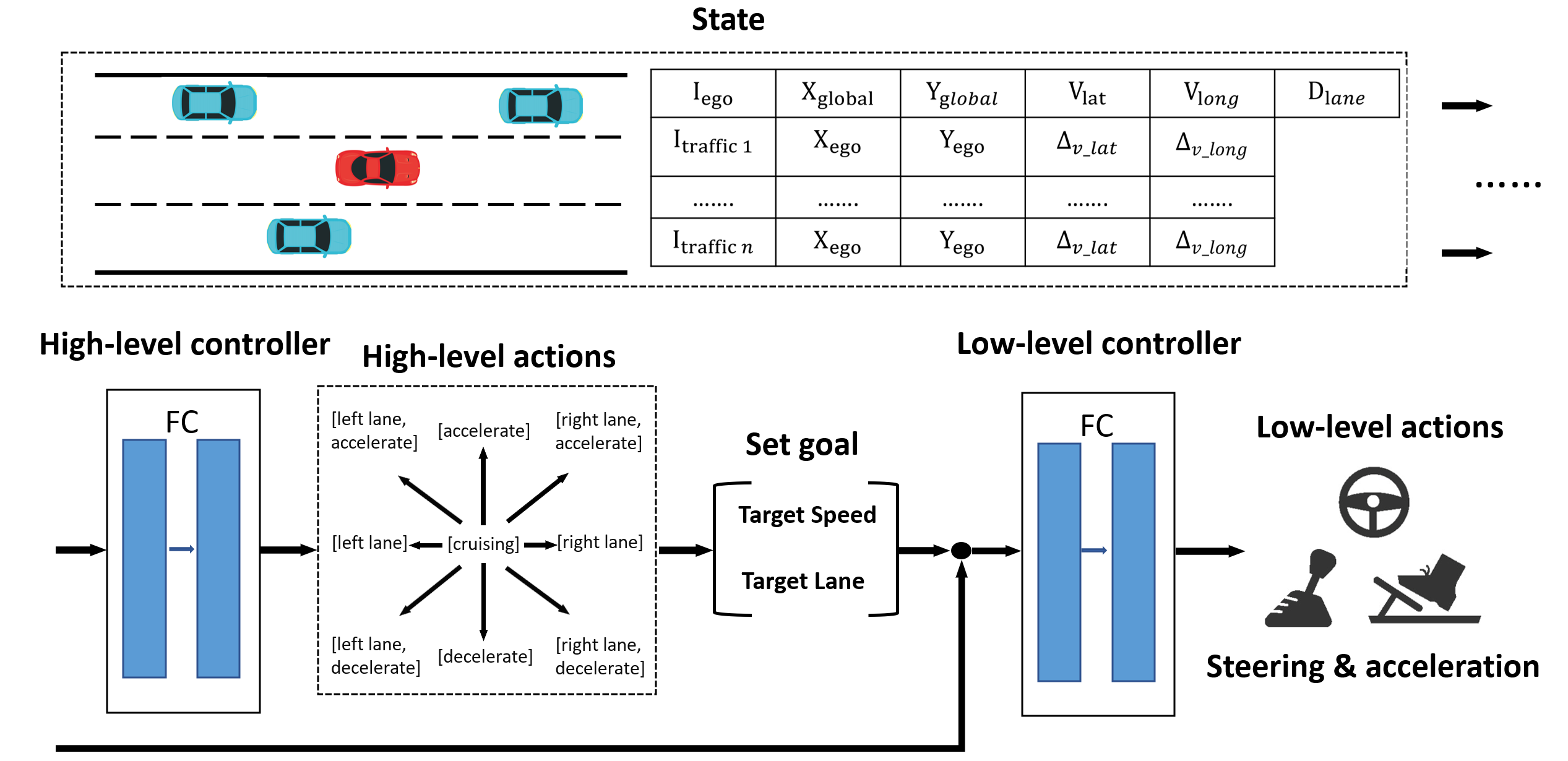}
\caption{Hierarchical DRL framework for highway driving.}
\label{framework}
\end{figure*}

\section{Related work}
Automated driving systems typically encompass two primary components: environment modeling and vehicle control. These systems deploy an array of sensors, such as lidars and cameras, to observe the environment. Through data interpretation algorithms, these systems can detect, classify, and track relevant features and objects, while simultaneously pinpointing their own location within a dynamic driving scene\cite{yurtsever2020survey,kiran2021deep}.

Once the driving environment is adequately represented, the vehicle's autonomous agent engages in the planning and decision-making process. Hierarchical controllers introduce a layered architecture, decomposing complex decision-making into manageable layers in order to enhance system efficiency, manageability, and interpretability relative to single-level controllers. Often, the automated driving tasks are artificially segmented to simplify the overall driving challenge\cite{kurt2013hierarchical,rezaee2019multi,peng2022integrated,wang2018reinforcement,chen2018deep,wang2020learning,sonu2018exploiting,moghadam2019hierarchical,albrecht2021interpretable}. For instance, separate high-level controllers might be designed for distinct traffic scenarios or for each high-level decision. This hierarchical decomposition allows for targeted solutions to specific driving tasks, with the potential for model-based control methods at the lower levels to ensure lightweight and robust responses.

The common limitations associated with current hierarchical DRL controllers for autonomous vehicles can be encapsulated within three primary aspects:

1.	Oversimplified Testing Environments: The complexity of testing environments is significantly limited by factors such as homogeneous traffic types\cite{sonu2018exploiting}, sparse traffic volume\cite{wang2020learning,chen2018deep}, reduced vehicle speeds\cite{wang2020learning}, and fewer lanes\cite{xu2018reinforcement}. Another common limitation is a constrained vehicle action space such that it can engage in either a lateral or longitudinal control action, but not both simultaneously\cite{peng2022integrated}. These constraints do not accurately reflect the intricate and variable conditions typical of real-world automated driving.

2.	Controller Design Constraints: 
For some hierarchical DRL controllers, the optimal driving solutions are primarily derived from the high-level controller. The low-level controller relies on rule-based and model-based\cite{moghadam2019hierarchical,shi2019driving} approaches. This design approach hampers the possibilities for optimizing low-level motion planning strategies that are adaptive to the current high-level commands. Consequently, there remains a necessity to design or train an optimal low-level controller for each specific high-level controller, which does not fully exploit the potential of reinforcement learning for vehicular control.

3. Environmental Exploration Capabilities: Current implementations do not fully enable the hierarchical DRL controller's ability to improve the exploration of environments. Integrating the high-level controller for long-term strategy and the low-level controller for immediate actions can improve the system's performance and adaptability in complex scenarios.
Huilkarni et al.~\cite{kulkarni2016hierarchical} proposed a hierarchical framework that facilitates efficient exploration in complex environments, demonstrating its efficacy in an ATARI game with sparse feedback. This hierarchical DRL improves environment exploration by simultaneously training a high-level controller for goal setting and a low-level controller for action execution. 
However, transferring this framework to the context of highway driving presents unique challenges. The dynamic nature of highway traffic, with its continuously evolving scenarios, demands a more adaptable approach. Moreover, the distinct reward systems for high and low-level controllers necessitate careful consideration to ensure that the low-level controller's actions align with the overarching driving objectives.

In addressing these challenges, our work builds upon the hierarchical DRL paradigm, aiming to refine controller robustness and adaptability.

\section{Problem formulation}

In our methodology, we formulate the task of automated driving on highways as a Markov Decision Process (MDP) problem\cite{sutton2018reinforcement}. MDP  can be described by a tuple $\langle S, A, P_{ss^\prime}, R_s \rangle$ where the states \( S \) encapsulate the vehicle's environment and its internal status, while the actions \( A \) reflect the vehicle's potential maneuvers. The transition between state-actions is denoted as $P_{ss^\prime}=P\left[S_{t+1}=s^\prime\middle| S_t=s\right]$. The rewards $R_s=E\left[R_{t+1}\mid S_t=s\right]$ are designed to guide the vehicle towards optimal driving behavior. A DRL-based controller aims to optimize the cumulative reward within a single episode, which is derived from the external reward function.

\subsection{State}
\begin{figure*}[!t]
\centering
\includegraphics[width=\textwidth]{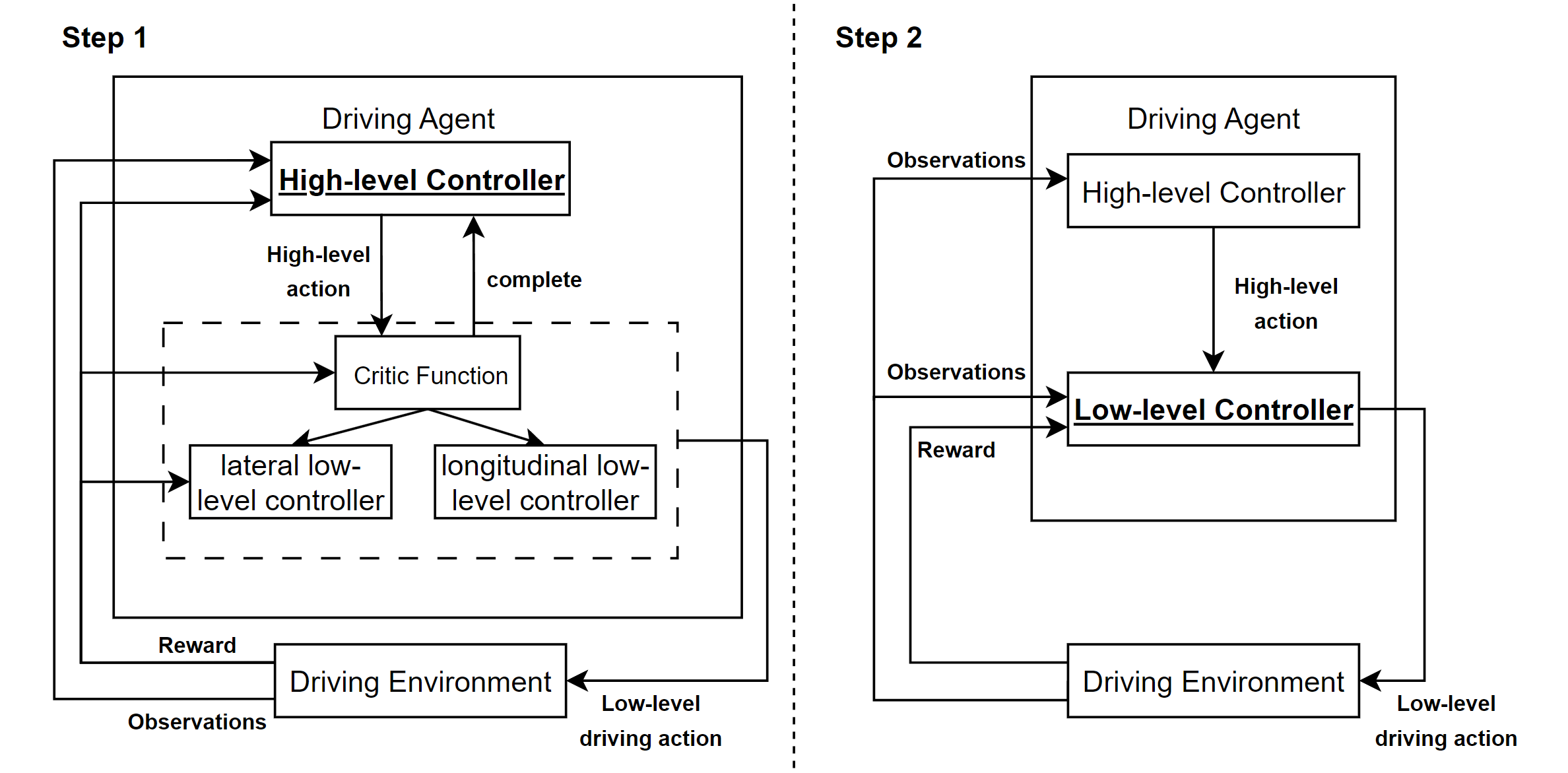}
\caption{Two-Step Training Process for High-Level and Low-Level Frameworks: Initially, the high-level controller is trained using a model-based motion planner and a critic function. Subsequently, this trained high-level controller is employed to facilitate the training of the low-level controller.}
\label{training framework}
\end{figure*}

The observation model in our study comprehensively captures the state information of the ego vehicle and its interaction with four surrounding vehicles, incorporating a total of 26 distinct factors. The state of the ego vehicle is characterized by several parameters. The presence of the ego vehicle is constantly indicated by a binary flag, $I_{ego}$, which in our scenario is always set to true, reflecting the ego vehicle's persistent presence. The ego vehicle's position in the global coordinate system is represented by $X_{global}$ and $Y_{global}$, denoting its longitudinal and lateral locations, respectively. Additionally, the ego vehicle's velocity is captured in terms of $V_{lat}$ and $V_{long}$, representing its lateral and longitudinal speeds in the global coordinate system. $d_{lane}$ represents the lateral distance of the ego vehicle from the nearest lane center. For each of the surrounding traffic vehicles, the state information includes a binary presence indicator, $I_{traffic}$, specifying whether the vehicle exists and is within the Region of Interest (ROI) of the ego vehicle.  $X_{ego}$ and $Y_{ego}$ represent the longitudinal and lateral distances, respectively, of the surrounding vehicles within the ROI from the ego vehicle. The lateral and longitudinal velocities of these vehicles relative to the ego vehicle's velocity are denoted by $\Delta v_{\text{\_lat}}$  and $\Delta v_{\text{\_long}}$.

\subsection{Action}
In this framework, at each timestep, the high-level controller processes the state information, \( S \), and selects a high-level action, $A_{high}$. The available high-level actions for the high-level controller are categorized into lateral and longitudinal decision-making processes. Lateral decisions include behaviors such as left lane change, maintaining the current lane, and right lane change. Longitudinal decisions involve adjustments to the vehicle's speed, either by increasing or decreasing it by a specified increment, \( \delta\) m/s. Lateral decisions do not interfere with longitudinal decisions. The selected high-level action is then interpreted as instructions for setting the target goal for the low-level controller, represented as \((l_{\text{target}}, V_{\text{target}})\). Here, \(l_{\text{target}}\) specifies a target lane index, and \(V_{\text{target}}\) designates a target speed. 

For the low-level DRL-based controller, we employ a discrete action space. The low-level action, $A_{low}$, consists of a steering and acceleration pair $({a},\theta)$.
The  acceleration $ a $ and the steering angle $\theta$ action spaces are defined as follows:
\begin{equation}
a \in\{-1, 0, 1\}m/s^2
\end{equation}
\begin{equation}
\theta \in\{-\pi/50,0,\pi/50\}rad
\end{equation}
Thus the automated driving agent operates with a total of 9 discrete low-level actions.

\subsection{Reward}

In our experiment settings, we define four primary reward terms: speed, lane centering, and steering rewards, along with a penalty for accidents. In the highway driving scenario, we mainly consider three types of accidents: the vehicle comes to a stop on the highway,  the vehicle oversteers and leaves the highway, and the vehicle collides with another vehicle. An episode concludes either when an accident occurs or when the maximum timesteps for an episode is reached.

\textbf{speed reward}:

\begin{equation}
\label{speedreward}
r_v = 
\begin{cases} 
e^{-(v-15)^2} & , v > 15 \\
\frac{8}{25}v - \frac{19}{5} & , 12.5 < v \leq 15 \\
\frac{2}{75}v - \frac{2}{15} & , 5 < v \leq 12.5 \\
0 & , v \leq 5 
\end{cases}
\end{equation}

The speed reward is structured around three distinct zones: the Dangerous Speed Zone ($v \in [0, 5] \, \text{m/s}$), the Low Speed Zone ($v \in (5, 12.5] \, \text{m/s}$), the Ideal Speed Zone ($v \in (12.5, 15] \, \text{m/s}$), and the Over Speed Zone ($v \in (15, \infty) \, \text{m/s}$). The Dangerous Speed Zone is characterized by speeds too low for highway safety, posing a risk of traffic accidents. The Low Speed Zone, while not hazardous, represents suboptimal vehicle speeds. Our target lies within the Ideal Speed Zone, with a specified speed of 15 m/s. If the ego vehicle's speed drops below this threshold, the controller encourages an acceleration back to 15 m/s. Conversely, the Over Speed Zone denotes speeds exceeding the limit, which are discouraged.

\textbf{lane-centering reward}:

\begin{equation}
r_y = e^{-1.5{\Delta d}^{2}}
\end{equation}

The lane-centering reward is designed to encourage the vehicle to maintain its position in the center of the lane, a critical factor for safe and efficient driving, especially on curves\cite{nageshrao2019autonomous}. $\Delta d $ represents the distance to the center of the current lane.

\textbf{steering reward}: 
\begin{equation}
\label{steering reward}
r_{\theta} = -|\sin(\theta)|
\end{equation}
The steering reward aims to prevent excessive steering, promote smoother driving, and reduce the risk of overcorrection.

\textbf{penalty for accident}: 

\begin{equation}
\label{total reward}
r_{\text{total}} = 
\begin{cases} 
\frac{w_v r_v + w_{\theta} r_{\theta} + w_y r_y}{w_v+w_{\theta}+w_y}
 & \text{,if no accident} \\
-10 & \text{,if accident}
\end{cases}
\end{equation}
Each of these reward terms is assigned a specific weight factor, which helps to balance their influence on the vehicle's learning process.  Additionally, the model imposes a significant penalty of -10 for critical failures such as crashes, driving out of the lane, or stopping in the lane. Specific weight values can be found in Table \ref{tab: Training parameters}.

\begin{figure}[!t]
\centering
\includegraphics[width=0.5\textwidth]{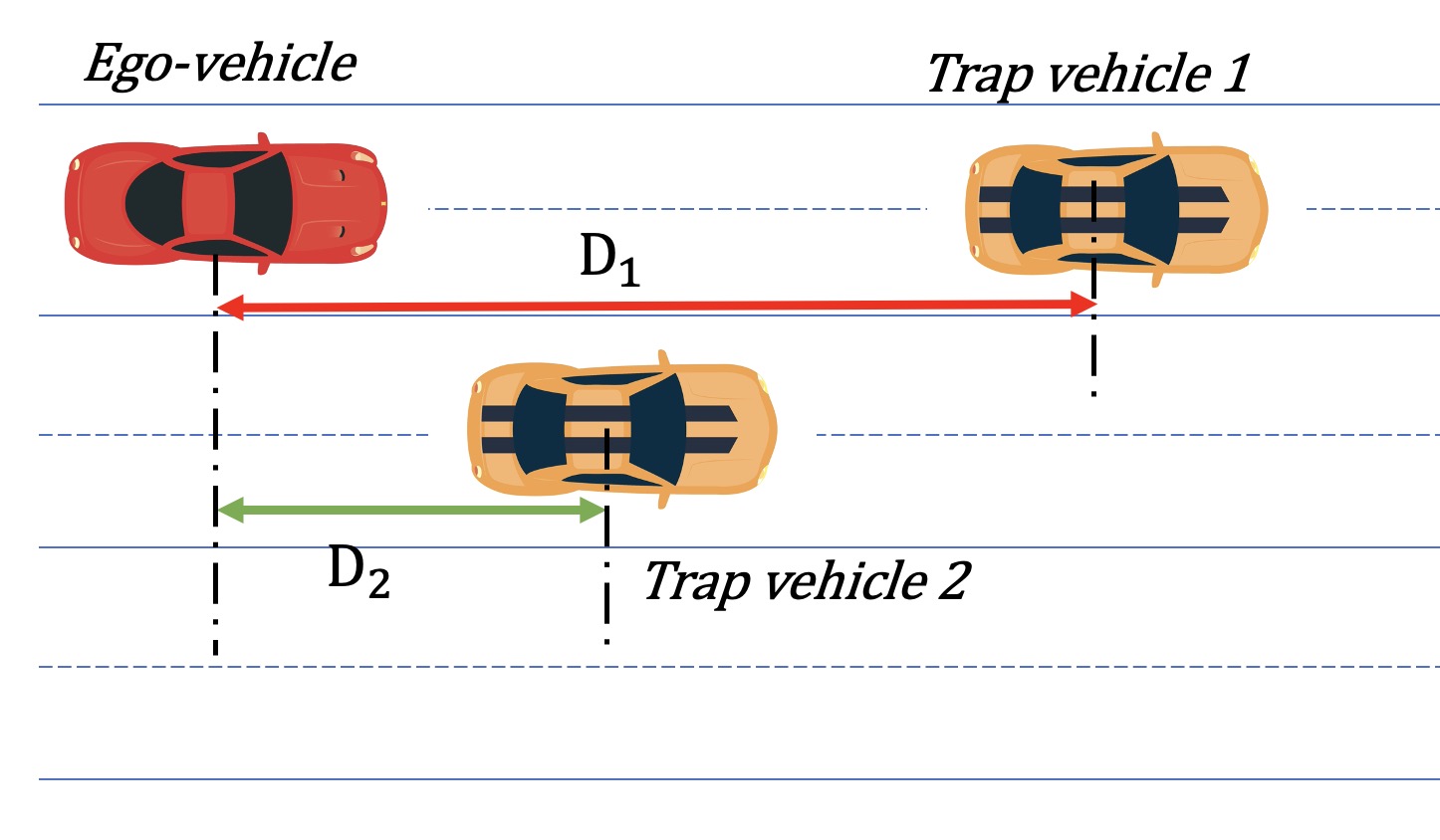}
\caption{The trap vehicles are initialized with a longitudinal distance with respect to the ego vehicle.}
\label{trapdefination}
\end{figure}
\subsection{Traffic vehicle control}

To control the traffic vehicles, we implement the Intelligent-Driver Model (IDM)\cite{treiber2000congested} for longitudinal control and the Minimizing Overall Braking Induced by Lane Change (MOBIL) model\cite{kesting2007general} for lateral decision-making. 

The IDM model\cite{treiber2000congested} calculates the desired acceleration of the vehicle using: 
\begin{equation}
\label{IDM equation}
acc = a \left[ 1 - \left( \frac{v}{v_{desired}} \right)^{\delta} - \left( \frac{s_{desired}}{s} \right)^2 \right]
\end{equation}
where ${s}$ is the gap to the front vehicle, $s_{desired}$ is the desired distance to the preceding vehicle, ${v}$ is the vehicle's current velocity, ${a}$ is a parameter for acceleration, and $v_{desired}$ is the target velocity for the controlled vehicle. 
The desired distance to the preceding vehicle is calculated using 
\begin{equation}
\label{IDM distance}
s_{desired} = S_0 + Tv + \frac{v \Delta v}{2 \sqrt{ab}}
\end{equation}
where ${S_0}$ is the desired distance gap, ${T}$ the desired time gap to the preceding vehicle, and ${a}$ and ${b}$ are acceleration and deceleration parameters. The desired distance is calculated based on the vehicle's speed and the relative speed between the current vehicle and the preceding vehicle.

The MOBIL model \cite{kesting2007general} decides when to perform a lane change based on the impact on other drivers: 
\begin{equation}
\label{second mobil constrain}
(a_e' - a_e) + p[(a_b' - a_b) + (a_a' - a_a)] > a_{\text{th}}
\end{equation}
where $(a_e' - a_e)$, $(a_b' - a_b)$, $(a_a' - a_a)$ represent the acceleration difference of the driver's vehicle, the following vehicle before the lane change and the following  vehicle after the lane change, respectively, $p$ represents the politeness factor which weighs the disadvantages imposed on other drivers due to the lane change, and $a_{\text{th}}$ is the acceleration threshold. The control parameters for the traffic vehicles are detailed in Table~\ref{tab: Traffic Vehicle parameters}.
\begin{table}[!ht]
\caption{Traffic vehicle parameters\label{tab: Traffic Vehicle parameters}}
\centering
\begin{tabular}{|c|c|} 
\hline
Parameters & value  \\
\hline
Vehicle length    & $ 5m$  \\
\hline
Vehicle Width   & $ 2m$  \\
\hline
Road Width   & $ 4m$  \\
\hline
Steering range $a_t$  & $ [-1, 1]m/s^2$  \\
\hline
Acceleration range $\theta_t$  & $ [-\pi/36,\pi/36] rad $  \\
\hline
Politeness factor p   &  0.5    \\
\hline
Acceleration threshold $a_{\text{th}}$   &  $0.2m/s^2$    \\
\hline
Acceleration $a$   &  $0.5m/s^2$    \\
\hline
Deceleration ${b}$ & $0.5m/s^2$ \\
\hline
Exponent value ${\delta}$   &  $4$    \\
\hline
Desired distance $s_{desired}$ & $10m$   \\
\hline
Desired velocity $v_{desired}$  & $12.5m$ \\
\hline
Desired distance gap ${S_0}$ & $10m$ \\
\hline
Desired time gap ${T}$ & $1.5s$ \\
\hline
\end{tabular}
\end{table}

\subsection{Trap initialization}
As previously mentioned, we define a typical trap scenario where the ego vehicle is in the far-left lane, with the trap vehicles positioned in front and in the lane to the right of the ego vehicle as shown in Figure~\ref{trapdefination}. ``Trap Vehicle 1'' is initialized with a longitudinal distance \( D_1 \) from the ego vehicle. ``Trap Vehicle 2'' maintains a close longitudinal distance \( D_2 \) from the ego vehicle, ensuring that it will block the ego vehicle if it attempts to change lanes or accelerate while maintaining its current speed. Both trap vehicles are set to travel at a speed slower than the normal traffic flow. To maintain the presence of this trap, we set both trap vehicles to the same speed for the entire episode. The initial distances between the ego vehicle and the trap vehicles \( D_1 \) and \( D_2 \) vary during training and testing. During training, we initial distances are drawn from a uniform distribution to introduce variability. For testing, we use more relaxed conditions. Detailed information can be found in Table~\ref{tab: Training parameters}.

\subsection{Deep reinforcement learning}

The core of the DDQN's\cite{van2016deep,mnih2015human} efficacy lies in its Q-value function. The Q-value function, denoted as \( Q(s, a) \), estimates the expected utility of taking action \( a \) in state \( s \), and then following the optimal policy thereafter. Formally, the Q-value function in the context of the DDQN is defined as follows:
%

\begin{equation}
\label{Q function}
\begin{aligned}
L(\theta) = \\\mathbb{E} \Bigg[ &\Bigg( r_t + \gamma \max_{a'} Q(s'_t, a'; \theta^-)- Q(s_t, a_t; \theta) \Bigg)^2 \Bigg]
\end{aligned}
\end{equation}
where \( r_t \) is the  reward for taking action \( a_t \) in state \( s_t \). \( \gamma \) is the discount factor. The loss is computed as the mean squared error (MSE) of the difference between the target Q-value (computed with \( \theta^- \)) and the predicted Q-value (computed with \( \theta \)) for the current state-action pair.

\section{Proposed method} 
\subsection{Hierarchical DRL framework}

Inspired by h-DQN \cite{kulkarni2016hierarchical}, we introduce our hierarchical DRL framework, as depicted in Figure \ref{framework}. The high-level controller is defined with an input layer corresponding to the number of observation features and an output layer corresponding to the number of goal features. It consists of two fully connected layers, each with 512 neurons, utilizing ReLU activations. The low-level controller takes the goal and new observations as inputs and outputs the discrete action pair. In practice, the goals will be interpreted as the difference in distance of the current state to the target lane center, $\Delta d_{\text{\_target}}$, and the difference between the current longitudinal speed and the target longitudinal speed,$\Delta v_{\text{\_target}}$. These can be represented by the following equations:
\begin{equation}
\Delta d_{\text{\_target}} = X_{\text{target\_lane}} - X_{\text{global}}
\end{equation}
\begin{equation}
\Delta v_{\text{\_target}} = V_{\text{target}} - V_{\text{long}}
\end{equation}
The low-level controller has the same structure as the high-level controller. The low-level action is the discrete steering and acceleration pair \((a, \theta)\) mentioned previously. During testing, at each timestep, the high-level controller first updates its goal based on the new observations. The low-level controller then combines this new goal with the new observations to generate the low-level actions within the same timestep. It is important to note that, unlike the h-DQN \cite{kulkarni2016hierarchical} design, where the performance of the low-level controller is critiqued by the intrinsic reward \cite{barto2013intrinsic}, the vehicle's performance in our framework is exclusively dependent on the external environment reward, which aligns with the high-level controller's training reward.

\subsection{Training the hierarchical DRL agent}

To separate the exploration ability of the high-level and low-level controller, we implement a two-step training method for the architecture shown in Figure~\ref{training framework}. The initial step only involves training the high-level controller. To facilitate using the high-level controller to control the vehicle exclusively, in this first step we use a rule-based motion planner to control the vehicle’s steering and acceleration based on the high-level goal. This also allows us to concentrate on the training and optimization of the high-level controller without the complexities introduced by an untrained low-level controller. To be specific, once the high-level controller sets the goal, the rule-based motion planner uses low-level action within the defined action space to reach the goal.  To evaluate whether the goals set by the high-level controller are being effectively reached, we introduce a critic function. If the lateral distance between the ego vehicle and the target lane center is less than a threshold $D_{\delta}$, the vehicle has successfully reached the target lane. If the speed difference between the current speed and target speed is less than a threshold $V_{\delta}$, the vehicle has reached the target speed. The lateral and longitudinal critic functions are thus:
\begin{equation}
\left| \Delta d_{\text{\_target}}\right| < D_{\delta}
\end{equation}
\begin{equation}
\left| \Delta v_{\text{\_target}} \right| < V_{\delta}
\end{equation}

If both the lateral and the longitudinal critic functions are satisfied, this means the rule-based motion planner has fully executed the high-level goal. The high-level controller is permitted to set a subsequent goal only upon the achievement of the current goal. We train the high-level controller for 1000 episodes and store the controller that has the best mean rewards for 10 episodes.

Once the high-level controller is adequately trained, we proceed to the second step: training the low-level controller. In this phase, we use the previously trained high-level controller and focus on the low-level controller, which operates at a frequency of 2 Hz. All controllers are using Epsilon-Greedy exploration strategy. The exploration probability is reduced from an initial value 0.5 to a final value of 0.02 over 1000 steps. Other training parameters are detailed in Table \ref{tab: Training parameters}. This targeted training allows the low-level controller to learn how to accurately control the vehicle under the guidance of the already optimized high-level controller. 

\begin{table}[!ht]
\caption{Experiments parameters\label{tab: Training parameters}}
\centering
\begin{tabular}{|c|c|} 
\hline
Parameters & value  \\
\hline
speed reward weight $w_v$ & 1.5  \\
\hline
steering reward weight $w_{\theta}$& 0.05  \\
\hline
lane centering reward weight $w_y$ & 0.05  \\
\hline
threshold $D_{\delta}$ & 0.3 m  \\
\hline
threshold $V_{\delta}$ & 0.3 m/s  \\
\hline
training episode  & 2000  \\
\hline
Duration  & 250  \\
\hline
batch size & 64    \\
\hline
experience reply memories & 5E4   \\
\hline
learning rate & 1E-3 \\
\hline
discount & 0.8 \\
\hline
 ${D_1}$ during training & [14.80,16.44] m\\
\hline
 ${D_2}$ during training & [4.06,7.43] m\\
\hline
 ${D_1}$ during testing & 15.62 m\\
\hline
${D_2}$ during testing & 6.61 m\\
\hline
\end{tabular}
\end{table}

\begin{figure}[bt]
\centering
\subfloat[]{\includegraphics[width=0.5\textwidth]{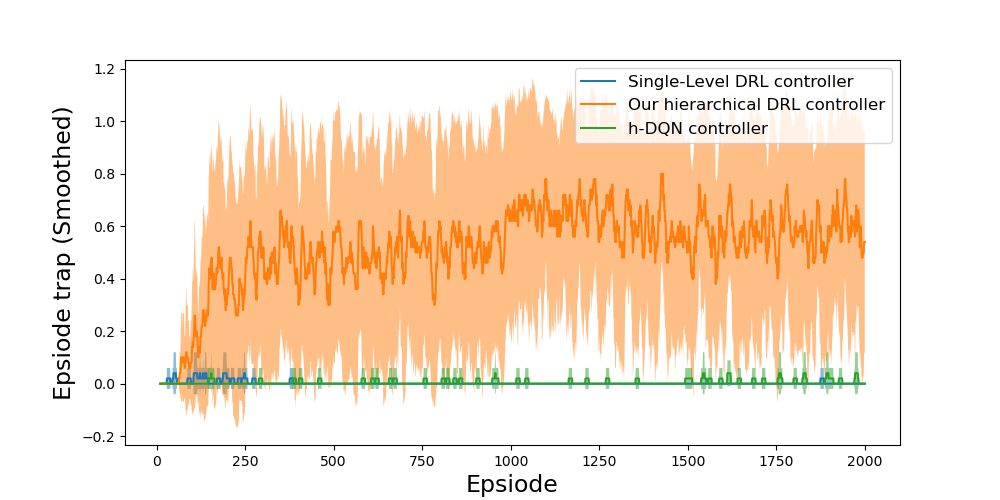}
\label{episodetrap}}
\\
\subfloat[]{\includegraphics[width=0.5\textwidth]{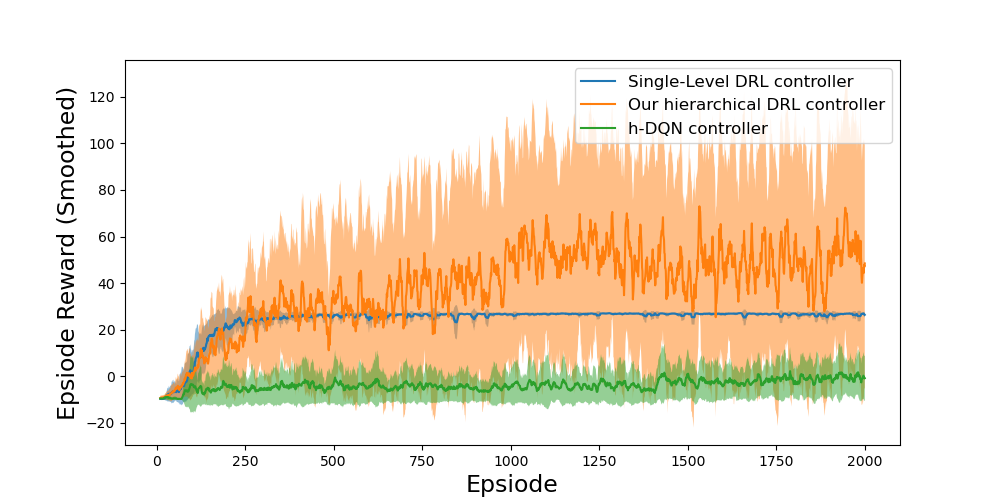}
\label{episodereward}}
\\
\subfloat[]{\includegraphics[width=0.5\textwidth]{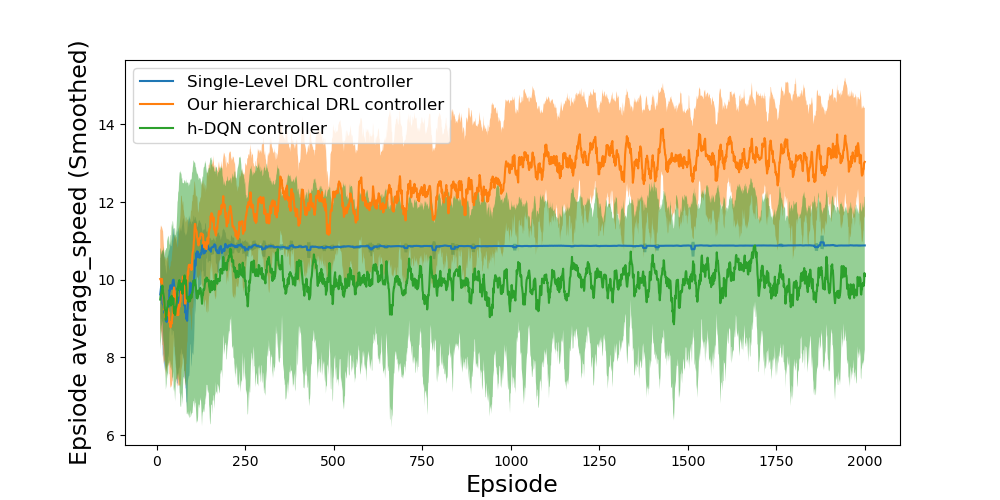}
\label{everagespeed}}
\caption{Single-level, hierarchical DRL, and h-DQN controller performance during training episodes. (a) Average success rate in evading low-speed traffic: the hierarchical DRL controller exhibits a higher likelihood of successfully navigating out of situations encumbered by low-speed traffic. (b) Average reward: The hierarchical DRL demonstrates superior performance, yielding higher average rewards per episode. (c) Average speed: The hierarchical DRL controller consistently achieves a higher average speed.}
\label{trainingresults3}
\end{figure}
\section{Experiment}
To test the algorithm's capability in environmental exploration while prioritizing long-term rewards, our approach involved integrating slow-moving vehicles as a constraint on the ego vehicle's maneuvers. Our experiments were conducted in the highway-env\cite{highway-env}. The criterion for successfully escaping a traffic trap is defined as the ego vehicle's ability to find an overtaking path within the episode duration. Specifically, the ego vehicle is deemed to have successfully escaped the trap if its rear bumper surpasses the front of all trap vehicles before the episode terminates. 
\begin{figure}[!t]
\centering
\subfloat[]{\includegraphics[width=0.5\textwidth]{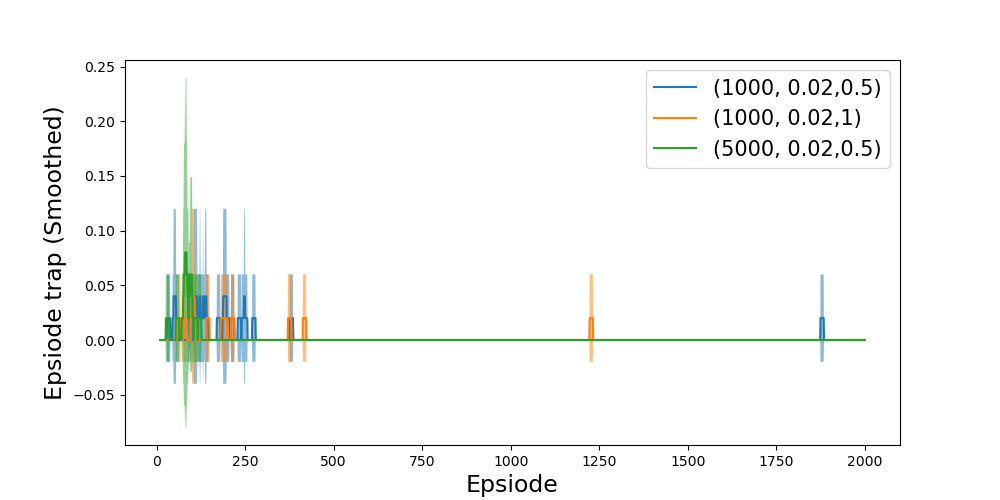}
\label{env_trap(explore)}}
\\
\subfloat[]{\includegraphics[width=0.5\textwidth]{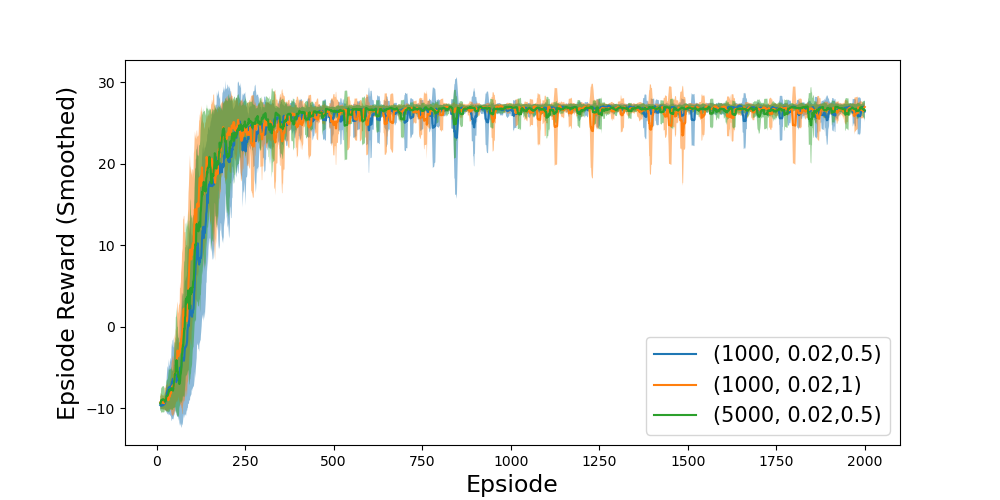}
\label{env_Reward(explore)}}
\\
\subfloat[]{\includegraphics[width=0.5\textwidth]{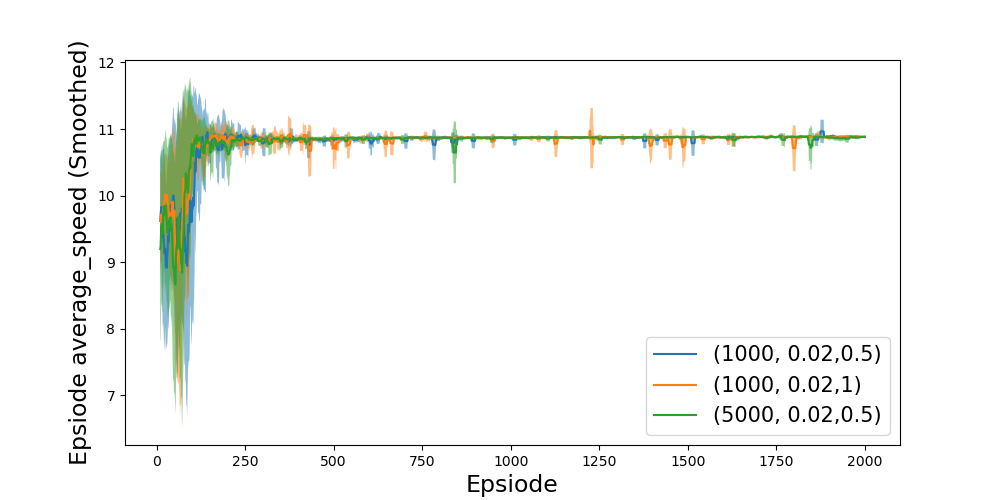}
\label{everage_speed(explore)}}
\caption{Comparing three single-level DRL controllers with different initial exploration. (a) The success rate in evading low-speed Traffic within a training episode. (b) Average reward within a training episode. (c) Average speed during training.} 
\label{trainingresultsSL}
\end{figure}
In our study, the single-level DRL controller employs a direct mapping from observation to optimal low-level action, utilizing two fully connected layers, each with 512 parameters, and is trained in a single step, in contrast to the hierarchical DRL controller which is trained using the two-step process described above.

\begin{figure*}[!t]
\centering
\includegraphics[width=\textwidth,height=0.4\textheight]{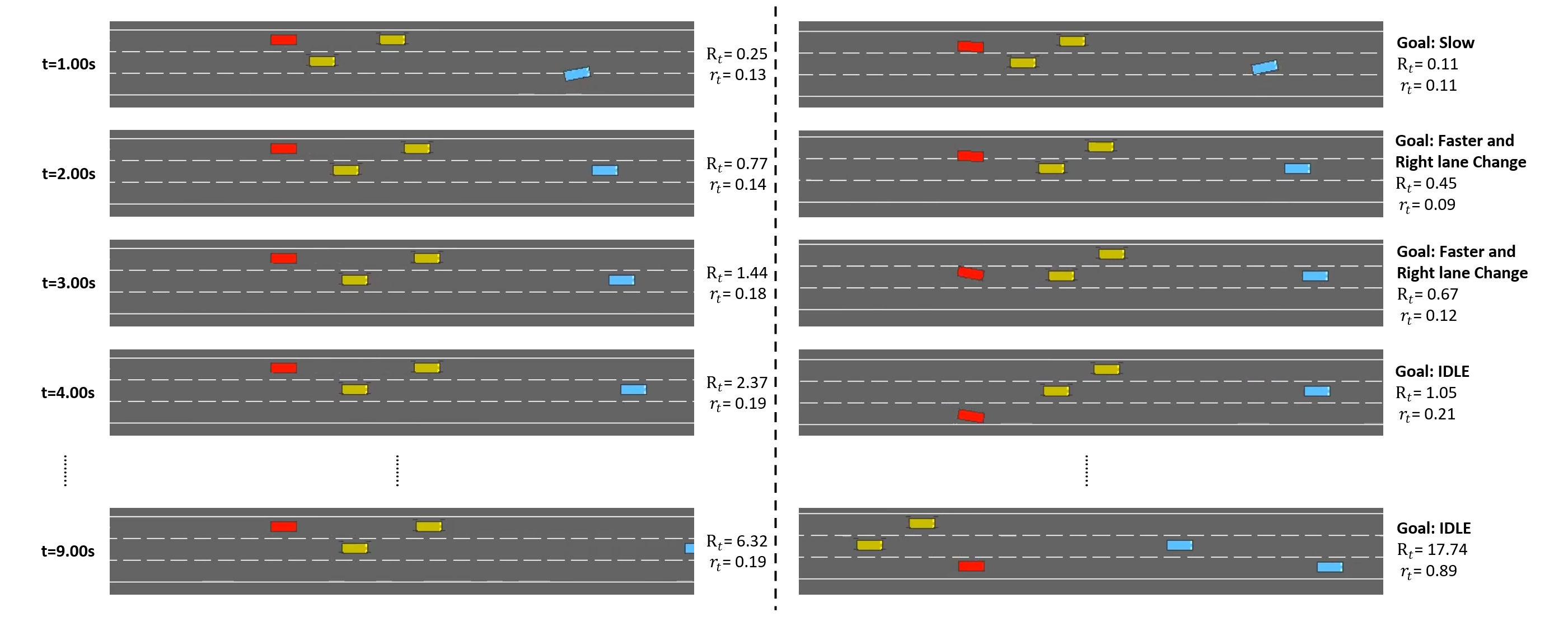}
\caption{The driving policies for slow-moving vehicles as implemented by both single-level and hierarchical DRL controllers. The red, yellow, and blue vehicles represent the ego, trap, and traffic vehicles, respectively. $r_{t} $ represents the instant reward received at video rendering time $t$. The accumulated reward $ R_t = \sum_{k=0}^{t} r_{t}$ represents the sum of instance rewards over time.} 
\label{video clip}
\end{figure*}

\begin{figure*}[!t]
\centering
\subfloat[]{\includegraphics[width=3.2in]{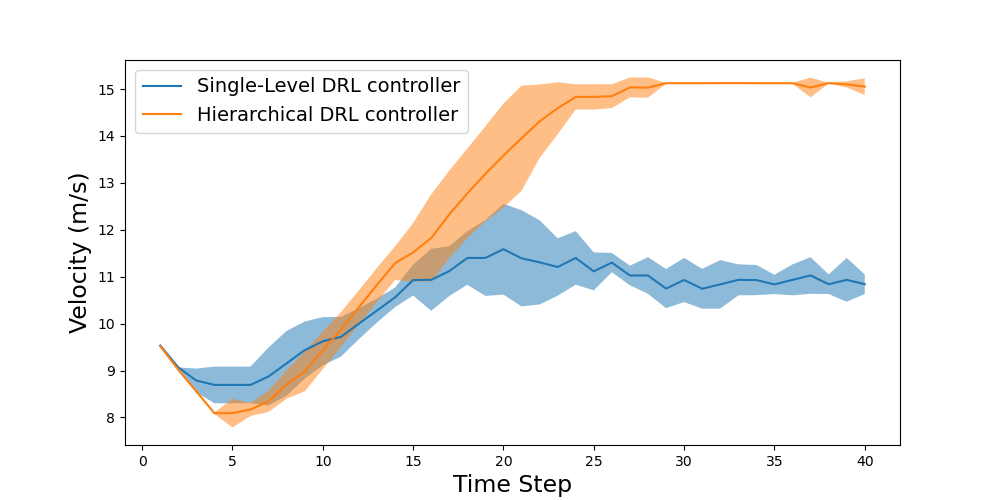}
\label{5velocity}}
\hfil
\subfloat[]{\includegraphics[width=3.2in]{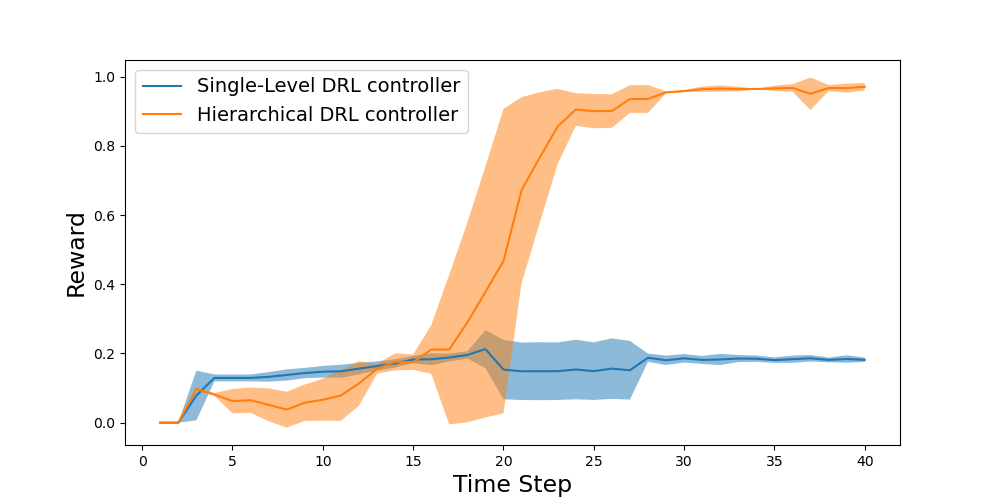}
\label{5Reward}}
\caption{(a) Velocity profile of the initial 40 timesteps of the single-level and hierarchical DRL controller. (b) Reward profile of the initial 40 timesteps of the single-level and hierarchical DRL controller.}
\label{profile}
\end{figure*}
We compared the training processes of the hierarchical DRL controller and the single-level DRL controller in three aspects: the average success rate of escaping traps (Figure \ref{episodetrap}), the average reward (Figure \ref{episodereward}), and the average speed (Figure \ref{everagespeed}) within an episode. Figure \ref{episodetrap} shows that during the initial exploration phase, both the hierarchical DRL controller and the single-level DRL controller exhibited tendencies to escape from traps. However, as the exploration probability decreased, the single-level DRL controller showed a minimal probability of escaping from difficult situations.

Figure \ref{everagespeed} and Figure \ref{episodereward} compare the differences in average speed and average reward per episode between the hierarchical DRL controller and the single-level DRL controller. Throughout the learning process, the single-level DRL controller's strategy revolved around maintaining speed and safety distance from the trap vehicle ahead. It failed to adopt overtaking strategies, limiting the ego vehicle's potential for long-term speed gains. Consequently, due to the adoption of a cruising strategy that matched the speed of the vehicle ahead, the ego vehicle achieved nearly constant average speed and reward in each episode. On the other hand, our hierarchical DRL controller enabled the agent to accelerate and closely approach the target speed of 15m/s after overtaking the trap vehicle, thereby obtaining a higher long-term speed reward.

It is noteworthy that the hierarchical DRL controller exhibited greater variance in all of the evaluation criteria compared to the single-level DRL controller. The single-level DRL controller has learned a relatively simple following strategy from a straightforward trap traffic pattern which has less uncertainty, whereas the hierarchical DRL controller had the additional challenge of learning overtaking maneuvers while subsequently adapting to the surrounding traffic.

To compare the exploration capability of the hierarchical controller in the environment, we enhanced the exploration ability of the single-level DRL controller using Epsilon-Greedy strategy to a certain extent. We refined the initial exploration strategy by tuning the epsilon parameter in the Epsilon-Greedy algorithm of a single-level DRL controller. Within the same setting, we executed five training experiments, adjusting the epsilon value to anneal from an initial value of 1 to a final value of 0.02 across 1000 steps, and from 0.5 to 0.02 over 5000 steps. Despite these modifications, Figures \ref{env_trap(explore)}, \ref{env_Reward(explore)}, and \ref{everage_speed(explore)} reveal that the success rate for evading traps showed no significant enhancement compared to the baseline, essentially remaining negligible after 2000 episodes. This result implies that simply increasing the initial exploration probability within a certain range does not guarantee an improved ability to detect viable escape paths. In contrast, the hierarchical DRL controller steadily improved its ability to escape traps during the early stages and continued to show enhancements even when the exploration value stabilized.

To elucidate the driving policies for managing slow-moving vehicles, as implemented by both single-level and hierarchical DRL controllers, we present in Figure~\ref{video clip} the vehicle's position changes in an ego vehicle centered point of view at selected video rendering moments. At moments t=1s, 2s, 3s and 4s, the single-level DRL controller demonstrated its ability to match the speed of the vehicle ahead while maintaining a safe distance. In contrast, the hierarchical DRL controller, guided by high-level objectives, initially reduced speed and then performed two lane changes to overtake the slower vehicle. This strategy led to a noticeable decrease in immediate rewards due to deceleration. By t=9s, after successfully navigating past the slow-moving vehicle, the hierarchical DRL controller achieved significantly higher immediate and accumulated rewards compared to the single-level DRL controller. Figures \ref{5velocity} and \ref{5Reward} further elucidate the driving policies over the initial 40 timesteps. The hierarchical DRL controller decreased its speed in the first 5 timesteps, then accelerated to approach the optimal speed. Although the single-level DRL controller also initially reduced speed, its objective was to maintain a safe following distance. Once this distance was secured, the single-level controller increased its speed to minimize the gap. The initial speed reduction by the hierarchical controller was more substantial than that of the single-level controller, necessitating the overcoming of the negative impact associated with prolonged periods of low reward. The hierarchical DRL controller decelerated and maintained a safe distance to execute a secure overtaking maneuver.

\begin{table}[!hb]
\caption{Training Phase Comparison\label{tab:training_comparison}}
\centering
\begin{tabular}{|m{2.5cm}|m{1.8cm}|m{1.8cm}|}
\hline
Training phase & Single-level DRL Controller & Hierarchical DRL Controller\\
\hline
Average reward & 25.82 & 47.94 \\
\hline
Average speed  & 10.89 & 13.089 \\
\hline
Average success rate to get out of trap  & 0\% & 60\%\\
\hline
\end{tabular}
\end{table}

\begin{table}[!ht]
\caption{Performance Comparison of Controllers in Overtaking Maneuver\label{tab:testing_comparison}}
\centering
\begin{tabular}{|m{2.5cm}|m{2cm}|m{2cm}|} 
\hline
Testing phase & Single-level DRL Controller & Hierarchical DRL controller \\
\hline
Average accumulated reward  & 3.32 &  13.20  \\
\hline
Average speed & 10.29  & 13.42  \\
\hline
Average success rate to get out of trap & 0\%  & 97.67\%  \\
\hline
Average travel distance & 257.29 & 331.60 \\
\hline
Accident rate & 0\% & 2.33\% \\
\hline
\end{tabular}
\end{table}

We further compare the average results for reward, speed, and the success rate of escaping traps over the final 50 episodes of the training period in 5 runs, as summarized in Table \ref{tab:training_comparison}. The hierarchical framework enhances the ego vehicle's ability to learn and execute escape maneuvers from traps through low-level actions.

After training the controllers, we conducted 300 episodes of tests in a relaxed trap scenario, each consisting of 25 timesteps. The The results of these tests are compiled in Table \ref{tab:testing_comparison}. In this consistent testing environment, the hierarchical DRL controller outperformed the single-level DRL controller in terms of average reward, average speed, and the average success rate of escaping traps, also covering a greater distance within the same time frame. However, due to the more conservative and simplistic driving strategy learned by the single-level DRL controller, it exhibited better performance in terms of the accident rate.

\section{Conclusion}
In this study, we have developed a hierarchical controller framework for autonomous highway driving to improve the robustness of handling complex traffic scenarios using DRL. We proposed using a high-level controller for long-term planning and broad exploration, while the low-level controller focuses on detailed vehicle maneuvering. To enhance the controller's effectiveness, we implemented a two-step training process, where each step focuses on training one controller.

We demonstrated the effectiveness of the hierarchical controller using a "Trap" situation, which is commonly seen in many highway scenarios. Based on our experiment, we proved that the hierarchical framework has a superior understanding of the "Trap" and therefore has a better chance to gain long-term rewards during training and testing episodes.
Future work could focus on utilizing more complex neural networks\cite{chen2021decision,janner2021offline} or incorporating safety constraints to ensure safer driving strategies while still being capable of handling complex traffic scenarios.

\bibliographystyle{unsrt}
\bibliography{ref}

\begin{thebibliography}{10}

\bibitem{kiran2021deep}
B~Ravi Kiran, Ibrahim Sobh, Victor Talpaert, Patrick Mannion, Ahmad~A Al~Sallab, Senthil Yogamani, and Patrick P{\'e}rez.
\newblock Deep reinforcement learning for autonomous driving: A survey.
\newblock {\em IEEE Transactions on Intelligent Transportation Systems}, 23(6):4909--4926, 2021.

\bibitem{yurtsever2020survey}
Ekim Yurtsever, Jacob Lambert, Alexander Carballo, and Kazuya Takeda.
\newblock A survey of autonomous driving: Common practices and emerging technologies.
\newblock {\em IEEE access}, 8:58443--58469, 2020.

\bibitem{schubert2011evaluating}
Robin Schubert.
\newblock Evaluating the utility of driving: Toward automated decision making under uncertainty.
\newblock {\em IEEE Transactions on Intelligent Transportation Systems}, 13(1):354--364, 2011.

\bibitem{varaiya1991sketch}
Pravin Varaiya and Steven~E Shladover.
\newblock Sketch of an ivhs systems architecture.
\newblock In {\em Vehicle Navigation and Information Systems Conference, 1991}, volume~2, pages 909--922. IEEE, 1991.

\bibitem{ngai2011multiple}
Daniel Chi~Kit Ngai and Nelson Hon~Ching Yung.
\newblock A multiple-goal reinforcement learning method for complex vehicle overtaking maneuvers.
\newblock {\em IEEE Transactions on Intelligent Transportation Systems}, 12(2):509--522, 2011.

\bibitem{xu2018reinforcement}
Xin Xu, Lei Zuo, Xin Li, Lilin Qian, Junkai Ren, and Zhenping Sun.
\newblock A reinforcement learning approach to autonomous decision making of intelligent vehicles on highways.
\newblock {\em IEEE Transactions on Systems, Man, and Cybernetics: Systems}, 50(10):3884--3897, 2018.

\bibitem{rosenzweig2015review}
Juan Rosenzweig and Michael Bartl.
\newblock A review and analysis of literature on autonomous driving.
\newblock {\em E-Journal Making-of Innovation}, pages 1--57, 2015.

\bibitem{peng2022integrated}
Jiankun Peng, Siyu Zhang, Yang Zhou, and Zhibin Li.
\newblock An integrated model for autonomous speed and lane change decision-making based on deep reinforcement learning.
\newblock {\em IEEE Transactions on Intelligent Transportation Systems}, 23(11):21848--21860, 2022.

\bibitem{bevly2016lane}
David Bevly, Xiaolong Cao, Mikhail Gordon, Guchan Ozbilgin, David Kari, Brently Nelson, Jonathan Woodruff, Matthew Barth, Chase Murray, Arda Kurt, et~al.
\newblock Lane change and merge maneuvers for connected and automated vehicles: A survey.
\newblock {\em IEEE Transactions on Intelligent Vehicles}, 1(1):105--120, 2016.

\bibitem{hatipoglu2003automated}
Cem Hatipoglu, Umit Ozguner, and Keith~A Redmill.
\newblock Automated lane change controller design.
\newblock {\em IEEE transactions on intelligent transportation systems}, 4(1):13--22, 2003.

\bibitem{chandler1958traffic}
Robert~E Chandler, Robert Herman, and Elliott~W Montroll.
\newblock Traffic dynamics: studies in car following.
\newblock {\em Operations research}, 6(2):165--184, 1958.

\bibitem{gipps1981behavioural}
Peter~G Gipps.
\newblock A behavioural car-following model for computer simulation.
\newblock {\em Transportation research part B: methodological}, 15(2):105--111, 1981.

\bibitem{treiber2000congested}
Martin Treiber, Ansgar Hennecke, and Dirk Helbing.
\newblock Congested traffic states in empirical observations and microscopic simulations.
\newblock {\em Physical review E}, 62(2):1805, 2000.

\bibitem{kesting2007general}
Arne Kesting, Martin Treiber, and Dirk Helbing.
\newblock General lane-changing model mobil for car-following models.
\newblock {\em Transportation Research Record}, 1999(1):86--94, 2007.

\bibitem{han2019driving}
Teawon Han, Junbo Jing, and {\"U}mit {\"O}zg{\"u}ner.
\newblock Driving intention recognition and lane change prediction on the highway.
\newblock In {\em 2019 IEEE Intelligent Vehicles Symposium (IV)}, pages 957--962. IEEE, 2019.

\bibitem{bojarski2016end}
Mariusz Bojarski, Davide Del~Testa, Daniel Dworakowski, Bernhard Firner, Beat Flepp, Prasoon Goyal, Lawrence~D Jackel, Mathew Monfort, Urs Muller, Jiakai Zhang, et~al.
\newblock End to end learning for self-driving cars.
\newblock {\em arXiv preprint arXiv:1604.07316}, 2016.

\bibitem{bojarski2017explaining}
Mariusz Bojarski, Philip Yeres, Anna Choromanska, Krzysztof Choromanski, Bernhard Firner, Lawrence Jackel, and Urs Muller.
\newblock Explaining how a deep neural network trained with end-to-end learning steers a car.
\newblock {\em arXiv preprint arXiv:1704.07911}, 2017.

\bibitem{codevilla2018end}
Felipe Codevilla, Matthias M{\"u}ller, Antonio L{\'o}pez, Vladlen Koltun, and Alexey Dosovitskiy.
\newblock End-to-end driving via conditional imitation learning.
\newblock In {\em 2018 IEEE international conference on robotics and automation (ICRA)}, pages 4693--4700. IEEE, 2018.

\bibitem{hawke2020urban}
Jeffrey Hawke, Richard Shen, Corina Gurau, Siddharth Sharma, Daniele Reda, Nikolay Nikolov, Przemys{\l}aw Mazur, Sean Micklethwaite, Nicolas Griffiths, Amar Shah, et~al.
\newblock Urban driving with conditional imitation learning.
\newblock In {\em 2020 IEEE International Conference on Robotics and Automation (ICRA)}, pages 251--257. IEEE, 2020.

\bibitem{yurtsever2020integrating}
Ekim Yurtsever, Linda Capito, Keith Redmill, and Umit Ozgune.
\newblock Integrating deep reinforcement learning with model-based path planners for automated driving.
\newblock In {\em 2020 IEEE Intelligent Vehicles Symposium (IV)}, pages 1311--1316. IEEE, 2020.

\bibitem{kendall2019learning}
Alex Kendall, Jeffrey Hawke, David Janz, Przemyslaw Mazur, Daniele Reda, John-Mark Allen, Vinh-Dieu Lam, Alex Bewley, and Amar Shah.
\newblock Learning to drive in a day.
\newblock In {\em 2019 International Conference on Robotics and Automation (ICRA)}, pages 8248--8254. IEEE, 2019.

\bibitem{chen2018deep}
Jianyu Chen, Zining Wang, and Masayoshi Tomizuka.
\newblock Deep hierarchical reinforcement learning for autonomous driving with distinct behaviors.
\newblock In {\em 2018 IEEE intelligent vehicles symposium (IV)}, pages 1239--1244. IEEE, 2018.

\bibitem{sonu2018exploiting}
Ekhlas Sonu, Zachary Sunberg, and Mykel~J Kochenderfer.
\newblock Exploiting hierarchy for scalable decision making in autonomous driving.
\newblock In {\em 2018 IEEE Intelligent Vehicles Symposium (IV)}, pages 2203--2208. IEEE, 2018.

\bibitem{kurt2013hierarchical}
Arda Kurt and {\"U}mit {\"O}zg{\"u}ner.
\newblock Hierarchical finite state machines for autonomous mobile systems.
\newblock {\em Control Engineering Practice}, 21(2):184--194, 2013.

\bibitem{rezaee2019multi}
Kasra Rezaee, Peyman Yadmellat, Masoud~S Nosrati, Elmira~Amirloo Abolfathi, Mohammed Elmahgiubi, and Jun Luo.
\newblock Multi-lane cruising using hierarchical planning and reinforcement learning.
\newblock In {\em 2019 IEEE Intelligent Transportation Systems Conference (ITSC)}, pages 1800--1806. IEEE, 2019.

\bibitem{wang2018reinforcement}
Pin Wang, Ching-Yao Chan, and Arnaud de~La~Fortelle.
\newblock A reinforcement learning based approach for automated lane change maneuvers.
\newblock In {\em 2018 IEEE Intelligent Vehicles Symposium (IV)}, pages 1379--1384. IEEE, 2018.

\bibitem{wang2020learning}
Jingke Wang, Yue Wang, Dongkun Zhang, Yezhou Yang, and Rong Xiong.
\newblock Learning hierarchical behavior and motion planning for autonomous driving.
\newblock In {\em 2020 IEEE/RSJ International Conference on Intelligent Robots and Systems (IROS)}, pages 2235--2242. IEEE, 2020.

\bibitem{moghadam2019hierarchical}
Majid Moghadam and Gabriel~Hugh Elkaim.
\newblock A hierarchical architecture for sequential decision-making in autonomous driving using deep reinforcement learning.
\newblock {\em arXiv preprint arXiv:1906.08464}, 2019.

\bibitem{albrecht2021interpretable}
Stefano~V Albrecht, Cillian Brewitt, John Wilhelm, Balint Gyevnar, Francisco Eiras, Mihai Dobre, and Subramanian Ramamoorthy.
\newblock Interpretable goal-based prediction and planning for autonomous driving.
\newblock In {\em 2021 IEEE International Conference on Robotics and Automation (ICRA)}, pages 1043--1049. IEEE, 2021.

\bibitem{shi2019driving}
Tianyu Shi, Pin Wang, Xuxin Cheng, Ching-Yao Chan, and Ding Huang.
\newblock Driving decision and control for automated lane change behavior based on deep reinforcement learning.
\newblock In {\em 2019 IEEE intelligent transportation systems conference (ITSC)}, pages 2895--2900. IEEE, 2019.

\bibitem{kulkarni2016hierarchical}
Tejas~D Kulkarni, Karthik Narasimhan, Ardavan Saeedi, and Josh Tenenbaum.
\newblock Hierarchical deep reinforcement learning: Integrating temporal abstraction and intrinsic motivation.
\newblock {\em Advances in neural information processing systems}, 29, 2016.

\bibitem{sutton2018reinforcement}
Richard~S Sutton and Andrew~G Barto.
\newblock {\em Reinforcement learning: An introduction}.
\newblock MIT press, 2018.

\bibitem{nageshrao2019autonomous}
Subramanya Nageshrao, H~Eric Tseng, and Dimitar Filev.
\newblock Autonomous highway driving using deep reinforcement learning.
\newblock In {\em 2019 IEEE International Conference on Systems, Man and Cybernetics (SMC)}, pages 2326--2331. IEEE, 2019.

\bibitem{van2016deep}
Hado Van~Hasselt, Arthur Guez, and David Silver.
\newblock Deep reinforcement learning with double q-learning.
\newblock In {\em Proceedings of the AAAI conference on artificial intelligence}, volume~30, 2016.

\bibitem{mnih2015human}
Volodymyr Mnih, Koray Kavukcuoglu, David Silver, Andrei~A Rusu, Joel Veness, Marc~G Bellemare, Alex Graves, Martin Riedmiller, Andreas~K Fidjeland, Georg Ostrovski, et~al.
\newblock Human-level control through deep reinforcement learning.
\newblock {\em nature}, 518(7540):529--533, 2015.

\bibitem{barto2013intrinsic}
Andrew~G Barto.
\newblock Intrinsic motivation and reinforcement learning.
\newblock {\em Intrinsically motivated learning in natural and artificial systems}, pages 17--47, 2013.

\bibitem{highway-env}
Edouard Leurent.
\newblock An environment for autonomous driving decision-making.
\newblock \url{https://github.com/eleurent/highway-env}, 2018.

\bibitem{chen2021decision}
Lili Chen, Kevin Lu, Aravind Rajeswaran, Kimin Lee, Aditya Grover, Misha Laskin, Pieter Abbeel, Aravind Srinivas, and Igor Mordatch.
\newblock Decision transformer: Reinforcement learning via sequence modeling.
\newblock {\em Advances in neural information processing systems}, 34:15084--15097, 2021.

\bibitem{janner2021offline}
Michael Janner, Qiyang Li, and Sergey Levine.
\newblock Offline reinforcement learning as one big sequence modeling problem.
\newblock {\em Advances in neural information processing systems}, 34:1273--1286, 2021.

\end{thebibliography}

\begin{IEEEbiography}[{\includegraphics[width=1in,height=1.25in,clip,keepaspectratio]{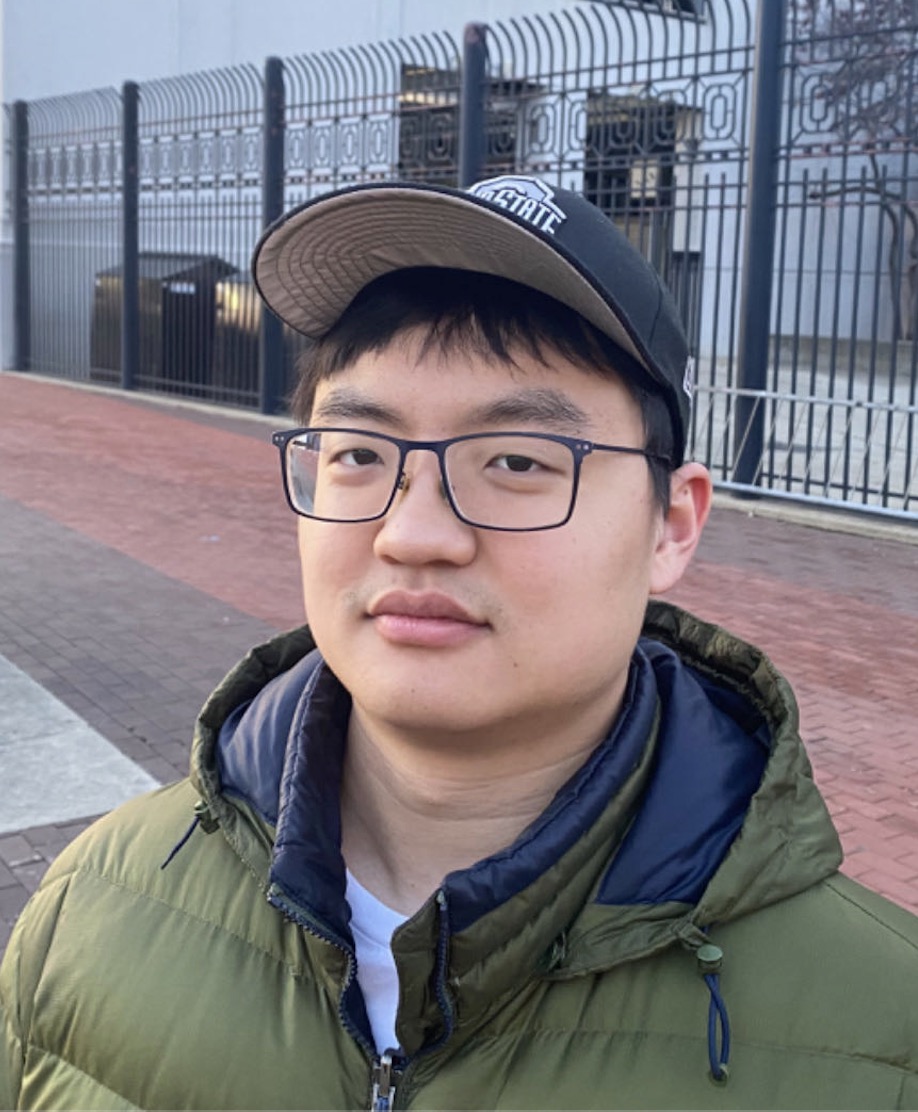}}]{Zhihao Zhang}
Zhihao Zhang recieved the M.S. degree in electrical and computer engineering from The Ohio State University, where he is currently pursuing the Ph.D. degree. His current research interest is automated control for vehicles.  
\end{IEEEbiography}
\begin{IEEEbiography}
[{\includegraphics[width=1in,height=1.25in,clip,keepaspectratio]{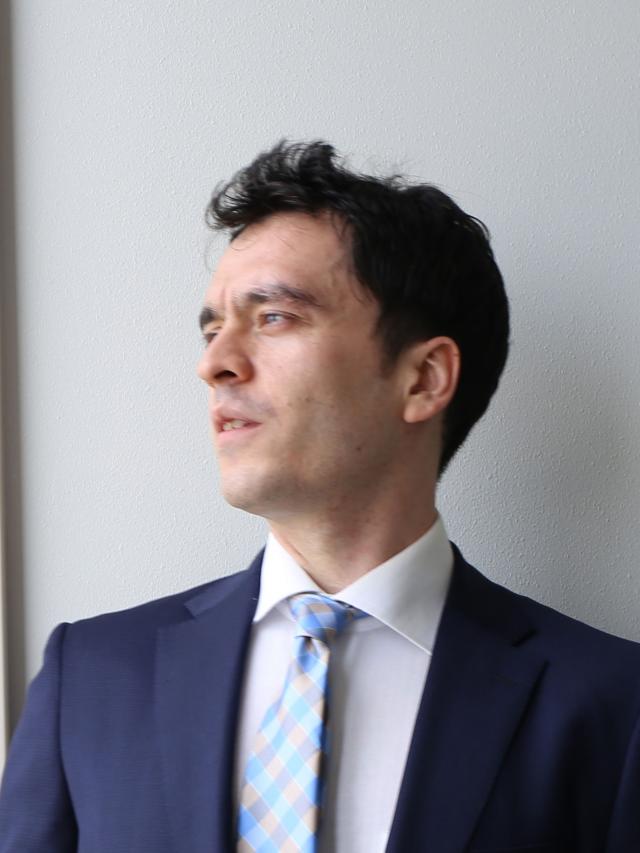}}]{Ekim Yurtsever}
 (Member, IEEE) received the B.S.
and M.S. degrees from Istanbul Tech. Uni., in 2012
and 2014, respectively, and the Ph.D. degree in
information science from Nagoya Uni., Japan, in
2019. Since 2019, he has been with the Department
of Elec. and Computer Engineering, The Ohio State
University, as a Research Associate. Currently, he is
working on bridging the gap between the advances
in the machine learning field and the intelligent
vehicle domain. His research interests include artificial intelligence, machine learning, computer vision,
reinforcement learning, intelligent transport. systems, and aut. driving systems.
\end{IEEEbiography}
\begin{IEEEbiography}
[{\includegraphics[width=1in,height=1.25in,clip,keepaspectratio]{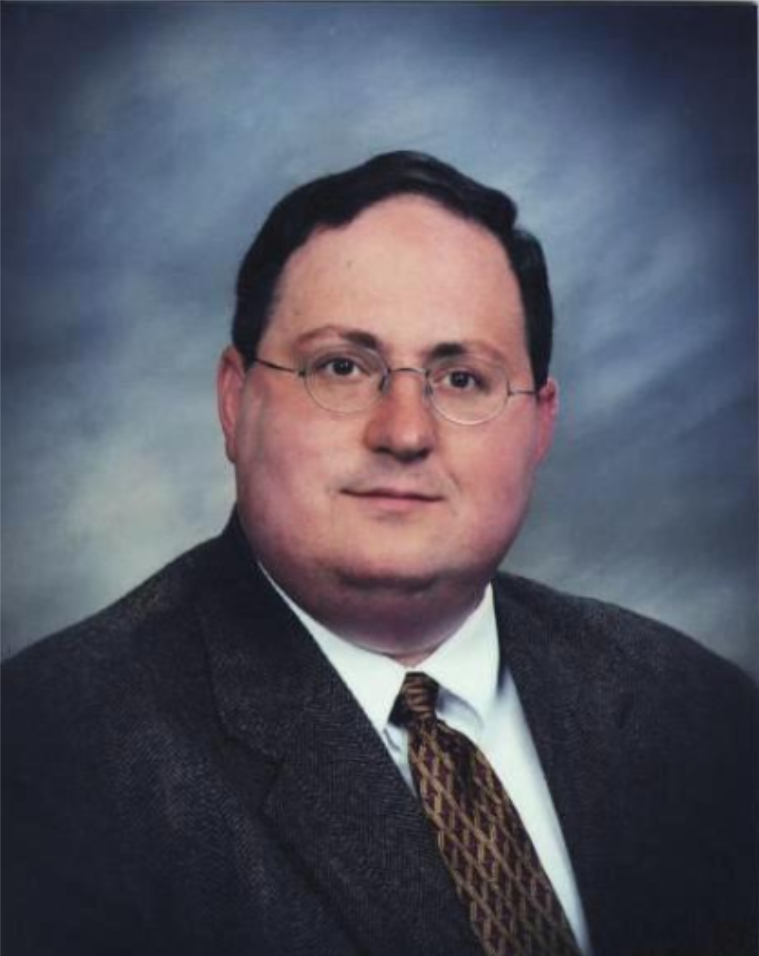}}]{Keith A. Redmill}
(Senior Member, IEEE) received the B.S.E.E. and B.A. degrees in mathematics from Duke University, Durham, NC, USA, in 1989, and the M.S. and Ph.D. degrees from The Ohio State University, Columbus, OH, USA, in 1991 and 1998, respectively. Since 1998, he has been with the Department of Electrical and Computer Engineering, The Ohio State University, initially as a Research Scientist. He is currently a Research Associate Professor. He is a coauthor of the book Autonomous Ground Vehicles. He has significant experience and expertise in intelligent transportation systems, intelligent vehicle control and safety systems, sensors and sensor fusion, wireless vehicle to vehicle communication, multiagent systems including autonomous ground and aerial vehicles and robots, systems, and control theory, virtual environment and dynamical systems modeling and simulator development, traffic monitoring and data collection, and real-time embedded and electromechanical systems.
\end{IEEEbiography}

\vfill
\end{document}